%% file: main.tex
\documentclass[runningheads]{llncs}
\usepackage[dvipsnames]{xcolor}
\usepackage[T1]{fontenc}
\usepackage{graphicx}
\usepackage{float}
\usepackage{comment}
\usepackage{graphicx}
\usepackage{amsmath, bm}
\usepackage{amssymb}
\usepackage{booktabs}
\usepackage{threeparttable,booktabs}
\usepackage{longtable}
\usepackage[hang,flushmargin]{footmisc}
\usepackage{multirow}
\usepackage{utfsym}
\usepackage{subfigure}
\usepackage{colortbl}
\usepackage{pgf}
\usepackage{tabularx}
\usepackage{adjustbox}
\usepackage{siunitx}
\usepackage{makecell}
\usepackage{array}
\usepackage[figuresright]{rotating}
\usepackage[pagebackref,breaklinks,colorlinks]{hyperref}
\definecolor{lightergray}{gray}{0.9}

\newcommand{\model}[0]{{\textcolor{black}{\textbf{RadIR}}}}

\colorlet{"0-7"}{GreenYellow}
\colorlet{"7-14"}{SpringGreen}
\colorlet{"14-"}{LimeGreen}
\colorlet{"0-2"}{GreenYellow}
\colorlet{"2-4"}{SpringGreen}
\colorlet{"4-"}{LimeGreen}
\definecolor{"bad"}{RGB}{246,161,147}

\begin{document}

\title{RadIR: A Scalable Framework for Multi-Grained Medical Image Retrieval via Radiology Report Mining}

\titlerunning{RadIR}

\author{Tengfei Zhang\inst{1,2}\textsuperscript{$*$} \and
Ziheng Zhao\inst{2,3}\textsuperscript{$*$} \and
Chaoyi Wu\inst{2,3} \and
Xiao Zhou\inst{2} \and \\
\vspace{.1cm}
Ya Zhang\inst{2,3} \and
Yanfeng Wang\inst{2,3}\textsuperscript{$\dag$} \and
Weidi Xie\inst{2,3}\textsuperscript{$\dag$}
}
\authorrunning {T. Zhang et al.}
\renewcommand{\thefootnote}{}
\footnotetext{$*$~Equal Contribution. ~ $\dag$~Corresponding Author}

\institute{
\setlength{\baselineskip}{14pt}
$^{1}$University of Science and Technology of China \\
$^{2}$Shanghai AI Laboratory \hspace{1cm}
$^{3}$Shanghai Jiao Tong University \\
\email{\{wangyanfeng622,weidi\}@sjtu.edu.cn}
}

\maketitle

\begin{abstract} 
Developing advanced medical imaging retrieval systems is challenging due to the varying definitions of `similar images' across different medical contexts. 
This challenge is compounded by the lack of large-scale, high-quality medical imaging retrieval datasets and benchmarks. In this paper, we propose a novel methodology that leverages dense radiology reports to define image-wise similarity ordering at multiple granularities in a scalable and fully automatic manner. Using this approach, we construct two comprehensive medical imaging retrieval datasets: \textbf{MIMIC-IR} for Chest X-rays and \textbf{CTRATE-IR} for CT scans, providing detailed image-image ranking annotations conditioned on diverse anatomical structures. Furthermore, we develop two retrieval systems, \textbf{RadIR-CXR} and \textbf{\model-ChestCT}, which demonstrate superior performance in traditional image-image and image-report retrieval tasks. These systems also enable flexible, effective image retrieval conditioned on specific anatomical structures described in text, achieving state-of-the-art results on 77 out of 78 metrics. 
\end{abstract}


\input{content/01_Introduction.tex}

\input{content/02_ProblemFormulation}

\input{content/03_DatasetCollection.tex}

\input{content/04_Method}

\input{content/05_Experiment.tex}

\input{content/06_Conclusion.tex}

\begin{credits}
\subsubsection{\ackname}
This study was funded by National Key R\&D Program of China (No. 2022ZD0161400).

\subsubsection{\discintname}
The authors have no competing interests to declare that are relevant to the content of this article.
\end{credits}

\bibliographystyle{splncs04}
\bibliography{egbib}

\clearpage
\newpage

\input{content/07_Appendix}

\end{document}

%% file: content/01_Introduction.tex
\section{Introduction}

The objective of this paper is to develop an image retrieval system for medical applications that ranks instances in a retrieval set based on their relevance to a query, which includes a radiology image and an optional text condition indicating the region to focus on, {\em i.e.}, the name of anatomy. Such a system has broad implications in enhancing clinicians’ ability to identify similar cases, supporting diagnosis and treatment planning, and facilitating medical education and research~\cite{choe2022content,qayyum2017medical,muller2004review,dubey2021decade}. Furthermore, in building generalist models~\cite{he2024meddr,RADFM,zhang2024development}, retrieval-augmented generation (RAG) plays a crucial role in reducing hallucinations and supporting case-based reasoning by grounding outputs in retrieved evidence.

Developing medical image retrieval systems is particularly challenging due to the complexity of defining image similarity, which depends on multiple factors such as global appearance, localized findings, and specific pathologies. 
For instance, two patients with different diseases may exhibit similar localized abnormalities. 
Capturing these nuanced relationships requires a granular understanding beyond coarse pathological or image-level labels. 
However, manual annotation of fine-grained similarity is often impractical due to its labor-intensive and subjective nature, especially at the scale needed for large datasets. Existing benchmarks~\cite{abacha20233d,denner2024leveraging,hu2022x,kobayashi2023sketch,lee2023region} typically rely on coarse image-level labels or limited manual annotations, which fail to capture the full spectrum of clinically relevant features, thereby limiting the development of scalable systems.

\begin{figure}[t]
\label{fig:teaser}
\centering
\includegraphics[width=1\textwidth]{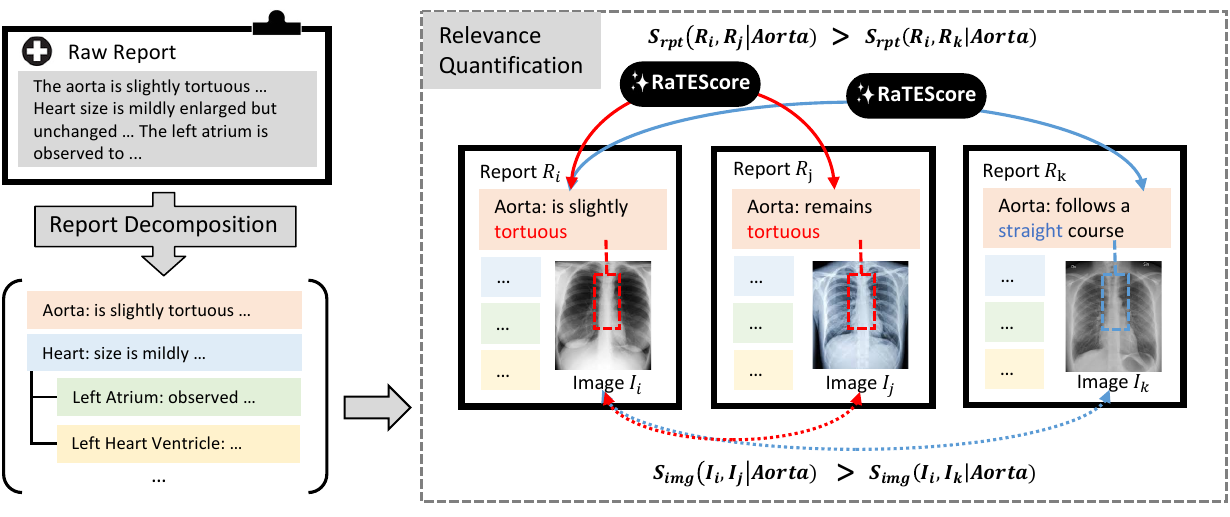}
\caption{\textbf{Fine-grained image similarity derived from report.} 
We decompose report into anatomy-centric findings and leveraging state-of-the-art medical language model RaTEScore to assess their relevance. We treat this as a proxy for the fine-grained image-image similarity, preserving their rankings in clinical meanings. }
\vspace{-0.2cm}
\end{figure}

To address the challenges 
we propose a novel medical image ranking pipeline by mining the multi-grained annotations from corresponding radiology reports. 
Specifically, given a certain anatomy structure, we first standardize the paired reports and extract the relative findings. Then, we adopt the text-level similarity ranking of the findings based on well-designed language-wise metrics~\cite{zhao2024ratescore}, to, in turn, represent the image similarity ranks regarding this anatomy structure. 
This pipeline enables the construction of multi-granular similarity ranking training data in a scalable and automated manner, for both global image matching and fine-grained retrieval conditioned on anatomy structures, as shown in Figure~\ref{fig:teaser}. 
Based on it, we extend two widely used datasets, MIMIC-CXR~\cite{johnson2019mimic} and CT-RATE~\cite{hamamci2024foundation}, to create two large-scale image retrieval datasets, \textbf{MIMIC-IR} and \textbf{CTRATE-IR}, with detailed annotations of image-image similarity ordering mined from dense report annotation, serving for both training and evaluation.

On model development, leveraging the two datasets, we have trained two retrieval systems: \textbf{RadIR-CXR} and \textbf{RadIR-ChestCT}. These systems achieve state-of-the-art performance in traditional image and image-report retrieval, while further enabling fine-grained retrieval with anatomy terminology as text condition. They allow users to query specific anatomies, bridging the gap between global similarity and localized retrieval, thus better fitting clinical demands. 

In summary, our contributions are threefold:
(i) We propose a novel, fully automated pipeline to structure radiology reports and bridge multi-grained image-image relevance in a scalable manner.
(ii) We develop \textbf{MIMIC-IR} and \textbf{CTRATE-IR}, two large-scale and comprehensive datasets accompanied by evaluation benchmarks for Chest X-ray and Chest CT image retrieval, with detailed annotations capturing image-image similarity ordering based on regional findings.
(iii) We present two state-of-the-art image retrieval systems, \textbf{RadIR-CXR} and \textbf{RadIR-ChestCT}, which demonstrate superior performance in global image retrieval and substantial advancements in image retrieval conditioned on anatomies.

%% file: content/02_ProblemFormulation.tex
\section{Problem Formulation}
\label{sec:problem_formulation}

Considering a collection of radiology image-report pairs, 
denoted as $\mathcal{D} = \{(I_1, R_1), \allowbreak ..., (I_K, R_K)\}$, 
where $I_i \in \mathbb{R}^{H\times W\times C}$ refers to the radiology image, and $R_i$ is the corresponding clinical report. 
The goal of the image retrieval task is to find the similar cases from $\mathcal{D}$, given a query image $I_q$ and optionally, a conditional query $Q$ referring to an anatomical structure. This is equivalent to ranking the candidates in $\mathcal{D}$ based on their relevance to the query image:
\begin{equation}
    \{{r_1}, {r_2}, ..., {r_K}\} = \mathcal{I}(\mathcal{S}_{\text{img}}(I_q, I_j \mid Q)),\text{ } \forall I_j \in \mathcal{D}
\end{equation}
where $r_i$ denotes the rankings, and $\mathcal{I}(\cdot)$ is a function that indexes the image similarity $\mathcal{S}_{img}(\cdot)$. 
When $Q$ is not provided, this reduces to a conventional image retrieval task without any conditions. 


\vspace{3pt} 
\noindent \textbf{Discussion. } 
In this ranking task, estimating the exact similarity values between images is unnecessary. Instead, we focus on preserving the relative similarity ordering. 
In this paper, we make the assumption that radiology reports have faithfully captured the critical findings of their paired images. Consequently, the similarity ranking of images should align with the similarity ranking of their corresponding reports. Thus, we use the similarity between radiology reports, denoted as $\mathcal{F}_{\text{rpt}}(\cdot)$, as a feasible and practical proxy for image similarity:
\begin{equation}
\label{equ:ordering_consistency1}
    \mathcal{I}(\mathcal{S}_{\text{img}}(I_q, I_j \mid Q)) = \mathcal{I}(\mathcal{S}_{\text{rpt}}(R_q, R_j \mid Q)), \text{ } \forall I_j, R_j \in \mathcal{D}
\end{equation}
The following sections detail the procedure for quantifying similarity between reports and leveraging these rankings to train the image retrieval system.

%% file: content/03_DatasetCollection.tex
\section{Dataset Construction}

In this section, we propose an automatic pipeline to quantify image-to-image similarity ordering, via mining their paired radiology reports, as shown in Figure~\ref{fig:teaser}. 
We first introduce the data sources in Section~\ref{sec:data_source}. 
Then, we detail the two main procedures in the pipeline: report decomposition in Section~\ref{sec:report_decomposition} and relevance quantification in Section~\ref{sec:relevance_quantification}. 

\subsection{Data Sources}
\label{sec:data_source}
We utilize two widely used datasets: \textbf{MIMIC-CXR}~\cite{johnson2019mimic} is the largest chest X-ray dataset, containing \textbf{377,110} image-report pairs; while \textbf{CT-RATE}~\cite{hamamci2024foundation} is a large-scale chest CT dataset with \textbf{25,692} non-contrast CT volume-report pairs. These reports include detailed radiological findings and impressions, which are essential for defining clinically meaningful similarities between images.  

\subsection{Report Decomposition}
\label{sec:report_decomposition}

We describe our process for extracting and structuring anatomical regions and their associated findings from radiology reports. This involves building a comprehensive anatomy terminology set, extracting regional findings, and integrating hierarchical relationships between anatomical structures, 
as detailed below.


\vspace{2pt} 
\noindent \textbf{Anatomy Terminology Set.} 
We utilized RadGraph-XL~\cite{delbrouck-etal-2024-radgraph} to extract anatomical structures from radiology reports. A total of 90 high-frequency anatomical structures commonly referenced in radiology were identified. To ensure consistency, synonymous terms ({\em e.g.}, ``superior vena cava'' and ``SVC'') were unified. The anatomical structures were further organized into a hierarchical framework, capturing relationships between parent structures ({\em e.g.}, ``lungs'') and their substructures ({\em e.g.}, ``left lung'' and ``right lung'').

\vspace{2pt} 
\noindent \textbf{Regional Findings Extraction.} 
From the `Findings' section of the reports, we extract region-specific findings by segmenting the content into sentences with periods as delimiters, and linking each sentence to the anatomical structures it mentions based on the anatomy terminology set.


\vspace{2pt} 
\noindent \textbf{Hierarchical Structure Integration.} 
Relationships between anatomies, such as ``lungs'' and ``left lung'', are utilized to merge findings from substructures into their parent structures. This integration provides a comprehensive, multi-level representation of findings for each anatomical region.

\subsection{Relevance Quantification}
\label{sec:relevance_quantification}

After performing fine-grained report decomposition, we can further quantify the relevance between findings from different reports regarding the same anatomy, as a substitute for the corresponding fine-grained image similarity on it. Here, we apply RaTEScore~\cite{zhao2024ratescore}, a state-of-the-art model that provides a robust evaluation metric for radiology report texts similarity based on key entities, as a proxy:
\begin{equation}
\label{equ:ratescore}
    \mathcal{S}_{\text{rpt}}(R_q, R_r \mid Q) = \text{RaTEScore}(\mathcal{E}(R_q \mid Q), \mathcal{E}(R_r \mid Q) ) 
\end{equation}
where $\mathcal{E}$ denotes extracting regional findings from the raw reports regarding $Q$ heuristically with rule-based string matching.
For image retrieval without specific query conditions that $Q$ is empty,  $\mathcal{E}$ will return the original report that evaluates the similarity of entire reports as a substitute for the global image similarity.

\vspace{0.3cm} \noindent \textbf{Summary.} 
We extract \textbf{2,582,477} regional findings in total, covering \textbf{90} anatomical structures. We further quantify over \textbf{132 billion} fine-grained image-image relevance between them.
We name the two proposed large-scale and multi-granularity datasets as \textbf{MIMIC-IR} and \textbf{CTRATE-IR}, as the foundation to train and benchmark the radiology image retrieval systems.

%% file: content/04_Method.tex
\begin{figure}[t]
\centering
\includegraphics[width=1\textwidth]{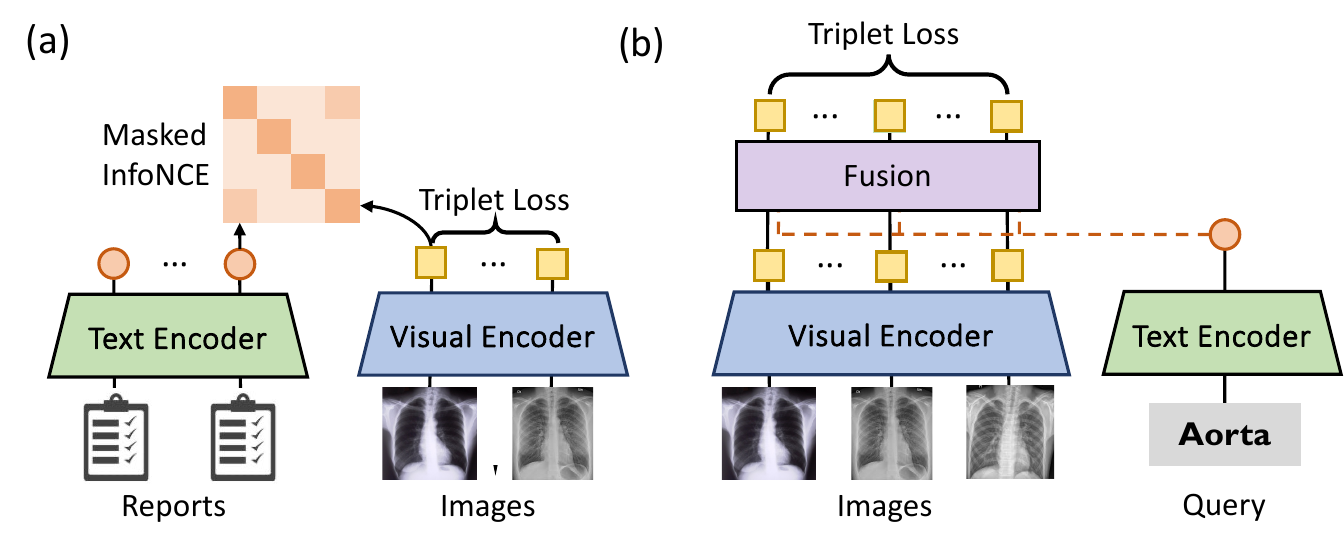}
\vspace{-.5cm}
\caption{\textbf{Architecture and training procedures of \model.} (a) In stage 1, we pre-train a CLIP-style model for unconditional image and image-report retrieval; (b) In stage 2, we extend the pre-trained model for image retrieval conditioned on anatomies.}
\label{img:arch}
\end{figure}

\section{\model}
In this section, we present the details to build \textbf{RadIR} based on the 
datasets we construct above. The training procedure includes two stages. In Section~\ref{sec:Overall_retrieval}, we pre-train the CLIP-style model for unconditional image retrieval; In Section~\ref{sec:Conditional_retrieval}, we extend the pre-trained model for retrieval task conditioned on a text query.

\subsection{Unconditional Image Retrieval}
\label{sec:Overall_retrieval}

\subsubsection{Architecture.} 
As shown in Figure~\ref{img:arch}(a), in this setting, we directly encode the raw images and reports without considering extra text queries.
We adopt a typical CLIP-style~\cite{radford2021learning} model with a Vision Transformer based image encoder $\Phi_{\text{visual}}(\cdot)$ and a BERT-based text encoder~$\Phi_{\text{text}}(\cdot)$:
\begin{equation}
v = \Phi_{\text{visual}}(I) \in \mathbb{R}^d, \quad t = \Phi_{\text{text}}(R) \in \mathbb{R}^d
\end{equation}
\noindent where $I$ denotes a radiology image, $R$ denotes a radiology report, $v$ and $t$ denotes their features respectively, and $d$ is the dimension.

\vspace{2pt}
\noindent \textbf{Training Objectives.}
Given a batch of $N$ samples, we can calculate the following similarity matrix as prediction: 
\begin{equation}
    \mathbf{{S}}_{i2t} = \boldsymbol{v}\boldsymbol{t}^T, \mathbf{{S}}_{t2i} = \boldsymbol{t}\boldsymbol{v}^T, \mathbf{{S}}_{i2i} = \boldsymbol{v}\boldsymbol{v}^T, \quad \boldsymbol{v} \text{ } \boldsymbol{t}\in \mathbb{R}^{N\times d}
\end{equation}
where $\mathbf{{S}}$ denotes the similarity matrices from image-text and image-image, respectively, and $\boldsymbol{v},\boldsymbol{t}$ denotes the visual or text embedding set.
Then, We applied masked infoNCE loss (MIL) ~\cite{infonceloss} and triplet loss (TL) ~\cite{Triplet_loss} to optimize our model:
\begin{equation}
    \mathcal{L} = \lambda_1 \mathcal{L}_\text{MIL}(\mathbf{{S}}_{i2t},\mathbf{T})+\lambda_2 \mathcal{L}_\text{MIL}({\mathbf{{S}}_{t2i}},\mathbf{T})+\lambda_3 \mathcal{L}_\text{TL}(\mathbf{S}_{i2i},\mathbf{T})
\end{equation}
where $\mathbf{T}\in \mathbb{R}^{N\times N}$ is a text-text similarity matrix calculated via RaTEScore, as illustrated in Section~\ref{equ:ratescore}. $\lambda_1,\lambda_2,\lambda_3$ are hyper-parameters. $\mathcal{L}_\text{MIL}$ is a variant of the classic infoNCE loss:
\begin{equation}
\mathcal{L}_{\text{MIL}}(\mathbf{S,T}) = -\frac{1}{N} \sum_{i=1}^{N} \log \left( \frac{\exp(\mathbf{S}_{ii})}{\sum_{j=1}^{N} \exp(\mathbf{S}_{ij}) \cdot \mathbf{(I+\mathbf{1}[\mathbf{T<\tau}])}_{ij}} \right)
\end{equation}
where $\mathbf{1}[\mathbf{T}<\tau]$ is a matrix that masks out the potential positive elements outside the diagonal, based on a predefined threshold $\tau$.
\input{content/tab/unconditional_results}

\subsection{Text-Conditioned Image Retrieval}
\label{sec:Conditional_retrieval}


\subsubsection{Architecture.} As shown in Figure~\ref{img:arch}(b), given  a conditional query $Q$, we employ a fusion module $\Phi_
\text{fusion}$ to extend the model for conditional image retrieval:
\begin{equation}
    f=\Phi_{\text{fusion}}(\Phi_{\text{visual}}(I),\Phi_{\text{text}}(Q)) \in R^d
\end{equation}
where $f$ denotes the fused feature. This enables the model to capture relevant visual features based on the anatomy.




\vspace{2pt}
\noindent \textbf{Training Objectives.} In contrast to the global similarity matrix $\mathbf{T}$ based on complete reports in Section~\ref{sec:Overall_retrieval}, we introduce $\mathbf{T}_{Q}$ as an anatomy-conditioned similarity matrix constructed from regional findings, based on equation~\ref{equ:ratescore}. Meanwhile, the predicted conditional image-image similarity result $\mathbf{S}_{f2f}$ is derived from the dot product of fused features. We then apply triplet loss on them: 
\begin{equation}
    \mathcal{L} = \mathcal{L}_\text{TL}(\mathbf{S}_{f2f},\mathbf{T}_{Q})
\end{equation}

%% file: content/tab/unconditional_results.tex
\begin{table}[t]
\centering
\caption{\textbf{Unconditional image to image, and image to report retrieval results.} Recall and NDCG results are presented in percentage. The best results on each metric are bolded.}
\vspace{-.1cm}
\resizebox*{0.85\textwidth}{!}{  
\footnotesize  
\setlength{\tabcolsep}{5pt}
\begin{tabular}{lcccccccc}
\toprule
\multirow{2}{*}{Method} & \multicolumn{4}{c}{$Recall@k \uparrow$} & \multicolumn{4}{c}{$NDCG\uparrow$} \\
\cmidrule(lr){2-5} \cmidrule(lr){6-9}
 & k=5 & k=10 & k=50 & k=100 & k=5 & k=10 & k=50 & k=100 \\
\midrule
& \multicolumn{8}{c}{\textit{on MIMIC-IR (Chest X-Ray)}} \\
\midrule
\multicolumn{9}{l}{\textbf{Image2Image}} \\
MedCLIP & 3.05 & 4.77 & 12.65 & 18.93 & 67.15 & 44.49 & 16.74 & 10.70 \\
BioMedCLIP & 2.04 & 3.30 & 8.20 & 12.68 & 64.49 & 42.72 & 16.07 & 10.27\\
PMC-CLIP &2.20 &3.58 &8.07 & 12.03 &63.23 & 41.88 & 15.75 &10.06 \\
RadIR-CXR & \textbf{5.18} & \textbf{6.94} & \textbf{15.45} & \textbf{21.29} & \textbf{68.23} & \textbf{45.21} & \textbf{17.01} & \textbf{10.88} \\
\midrule
\multicolumn{9}{l}{\textbf{Image2Text}} \\
MedCLIP & 0.19 & 0.28 & 2.04 & 3.77 & 58.22 & 38.58 & 14.51 & 9.27 \\
BioMedCLIP & 0.47 & 0.78 & 4.23 & 8.10 & 62.79 & 41.60 & 15.64 & 9.99 \\
PMC-CLIP & 0.31 & 0.44 & 2.73 & 5.43 & 50.96 & 35.18 & 13.99 & 9.06\\
RadIR-CXR & \textbf{4.33} & \textbf{6.88} & \textbf{18.18} & \textbf{25.34} & \textbf{69.07} & \textbf{45.76} &\textbf{ 17.21 }& \textbf{11.00} \\
\midrule
& \multicolumn{8}{c}{\textit{on CTRATE-IR (Chest CT)}} \\
\midrule
\multicolumn{9}{l}{\textbf{Image2Image}} \\
CT-CLIP & 19.43 & 28.76 & 57.51 & 68.13 & 74.48 & 75.20 & 78.05 & 79.96 \\
RadIR-ChestCT & \textbf{20.75} & \textbf{30.57} & \textbf{62.44} & \textbf{72.80} & \textbf{74.60} & \textbf{75.47} & \textbf{78.51 } & \textbf{80.49} \\
\midrule
\multicolumn{9}{l}{\textbf{Image2Text}} \\
CT-CLIP & 5.05 & 8.19 & 25.27 & 39.92 & 67.57 & 70.50 & 76.67 & 79.45 \\
RadIR-ChestCT & \textbf{6.65} & \textbf{12.99} & \textbf{36.72} & \textbf{52.91} & \textbf{69.18} & \textbf{72.11} & \textbf{78.12} & \textbf{80.84} \\
\bottomrule
\end{tabular}}  
\label{tab:similarity_test}
\vspace{-0.1cm}
\end{table}

%% file: content/05_Experiment.tex
\section{Experiment Settings and Results}

\input{content/tab/conditional_results}
\input{content/tab/conditional_results_CT}

We validate RadIR on MIMIC-IR and CTRATE-IR, with both unconditional image retrieval, image to report retrieval, and image retrieval conditioned on anatomy name. In all experiments, we follow the official train-test split of MIMIC-CXR and CT-RATE. In this section, we first introduce our baselines in Section~\ref{sec:baseline} and benchmark metrics in Section~\ref{sec:metric}; Then, we analyze the experiment results in Section~\ref{sec:global_retrieval} and Section~\ref{sec:conditional_retrieval}.

\subsection{Baseline}
\label{sec:baseline}

We take the following methods as baselines: \textbf{BioMedCLIP}~\cite{zhang2023biomedclip}, a vision-language foundation model for 2D biomedical images pre-trained on 15M image-text pairs; \textbf{MedCLIP}~\cite{wang2022medclip}, a decoupled image-text contrastive learning framework for chest X-Ray images trained on MIMIC-CXR~\cite{johnson2019mimic} and CheXpert~\cite{irvin2019chexpert}; \textbf{PMC-CLIP}~\cite{lin2023pmc}, a CLIP-style model pretrained on PMC-OA with 1.6M biomedical image-caption pairs; 
and \textbf{CT-CLIP}~\cite{hamamci2024foundation},a vision-language foundation model for Chest CT images pre-trained on CT-RATE~\cite{hamamci2024foundation}. Note that none of these baselines support conditional image retrieval, thus we evaluate their performance using retrieval results derived from holistic image and text features across all tasks.
 
\subsection{Metric}
\label{sec:metric}
\noindent \textbf{Recall@$k$} evaluates whether the correct items are in the top-$k$ predictions. In image-report retrieval, we consider the paired data as the correct item; In image-image retrieval, we view candidates with similarity over 0.9 as correct items.

\vspace{3pt}
\noindent \textbf{NDCG@$k$} evaluates the predicted ranking by comparing it with the ideal ranking. First, the Discounted Cumulative Gain (DCG) of a ranking is calculated as 
$\text{DCG} = \sum_{i=1}^{k} \dfrac{\text{rel}_i}{\log_2(i + 1)}$
, where $\text{rel}_i$ represents the ground-truth similarity score of the item ranked at position $i$, and $k$ is the number of items to consider in the ranking. The NDCG is defined as the ratio of the DCG of a predicted ranking to the DCG of the ideal ranking~(IDCG) obtained by sorting the items by ground-truth similarity score: 
$\text{NDCG} = \text{DCG}/\text{IDCG}$.

\subsection{Results on Unconditional Retrieval}
\label{sec:global_retrieval}

As demonstrated in Table~\ref{tab:similarity_test}, after fine-tuning, RadIR consistently exceeds the state-of-the-art CLIP models on image-image retrieval task, and on both CXR and Chest CT datasets. 
Notably, RadIR can also be applied for image-report retrieval and achieves notable improvement over baselines. 
These results highlight that RadIR can perform effectively in these traditional retrieval tasks.

\subsection{Results on Conditional Image Retrieval}
\label{sec:conditional_retrieval}

Table~\ref{tab:conditioanl_results_CXR} shows that RadIR outperforms baselines in 9 out of 10 anatomical regions on CXR images, and achieves the best performance on average; 
While in Table~\ref{tab:conditioanl_results_ChestCT}, RadIR consistently outperforms CT-CLIP on metrics. 
In addition, we observe that RadIR performs better on tail anatomies less frequently mentioned in the report. 
We hypothesize that this is because baselines trained on image-text pairs exhibit a bias towards more frequent anatomies. While RadIR, supporting conditional retrieval, can effectively adapt its focus to the queried anatomy, demonstrating superior robustness and versatility. 

%% file: content/tab/conditional_results.tex
\begin{table}[t]
\centering
\caption{\textbf{Conditional image retrieval results on MIMIC-IR.} Recall scores are averaged and aggregated by anatomical region, and presented in percentage. Anatomies are sorted in descending order of their frequency in train set, with the `head' regions at the top and the `tail' regions at the bottom. The best results for each anatomy are bolded. Greener suggests higher improvement over baselines.}
\label{tab:conditioanl_results_CXR}
\resizebox{\textwidth}{!}{%
\footnotesize
\begin{tabular}{@{}l *{12}{S[table-format=3.2]} @{}}
\toprule
\multirow{2}{*}{Anatomy} & \multicolumn{4}{c}{$Recall@3 \uparrow$} & \multicolumn{4}{c}{$Recall@5 \uparrow$} & \multicolumn{4}{c}{$Recall@10 \uparrow$} \\
\cmidrule(lr){2-5} \cmidrule(lr){6-9} \cmidrule(lr){10-13}
 & {PMC} & {BioMed} &{Med}& {RadIR} & {PMC} & {BioMed} &{Med}& {RadIR} & {PMC} & {BioMed} &{Med}& {RadIR} \\
 & {CLIP} & {CLIP} &{CLIP} & {} & {CLIP} &{CLIP}& {CLIP} & {} & {CLIP} &{CLIP}& {CLIP} & {} \\
\midrule
Pleura & 16.11 & 18.67 & 21.98 & \cellcolor{"2-4"} \textbf{25.32} & 23.51 & 28.35 & 30.44 & \cellcolor{"2-4"} \textbf{32.99} & 34.09 & 40.57 & 40.52 & \textbf{44.60} \cellcolor{"4-"}\\
Bones &  12.06 & 17.24 & 13.43 & \textbf{18.79} \cellcolor{"0-2"}& 20.97 & 24.74 & 21.91 & \textbf{26.16} \cellcolor{"0-2"} & 34.85 & 35.04 & 35.23 & \textbf{38.09} \cellcolor{"2-4"} \\
Lung & 7.08 & 9.48 & 8.62 & \textbf{11.37} \cellcolor{"2-4"} & 11.55 & 12.54 & 14.88 & \textbf{16.46} \cellcolor{"0-2"} & 18.03 & 19.03 & \textbf{22.41} & \cellcolor{"bad"} 22.31\\
Diaphragm & 16.11 & 17.75 & 18.95 &\textbf{21.13} \cellcolor{"2-4"} & 23.51 & 24.29 & 23.18 & \textbf{27.13} \cellcolor{"2-4"} & 34.09 & 36.65 & 34.95 & \textbf{37.77} \cellcolor{"0-2"}\\
Vascular & 14.80 & 19.57 & 22.93&\textbf{30.65} \cellcolor{"4-"}& 29.04 & 27.97 & 32.63 &\textbf{36.39}\cellcolor{"2-4"} &  39.96 & 39.85 & 45.76 & \textbf{47.37} \cellcolor{"0-2"} \\
Thorax & 4.24 & 11.84 & 8.61&\cellcolor{"2-4"} \textbf{14.52} & 12.96 & 17.02 &15.96 & \textbf{19.59} \cellcolor{"2-4"}& 20.83 & 25.42 & 27.37 & \textbf{29.27} \cellcolor{"0-2"}  \\
Heart & 11.51 & 10.44 & 7.90 & \textbf{16.02} \cellcolor{"2-4"} & 15.84 & 16.23 & 12.90 & \textbf{23.40} \cellcolor{"4-"}& 26.15 & 25.25 & 25.76 & \cellcolor{"4-"} \textbf{32.79}\\
Airway & 15.93 & 12.28 & 14.85 & \textbf{ 22.50} \cellcolor{"4-"}& 24.47 & 16.91  &  22.71 & \textbf{29.89} \cellcolor{"4-"}& 35.10 & 26.43 & 34.39 & \textbf{42.97} \cellcolor{"4-"} \\
Stomach & 7.41 & 11.85 & 12.59& \textbf{19.23} \cellcolor{"4-"} & 10.37 & 15.56 & 20.00 & \textbf{22.22} \cellcolor{"2-4"}& 22.96 & 23.70 & 24.44 & \textbf{31.11}  \cellcolor{"4-"}\\
Bronchi & 16.83 & 13.86 & 13.86 &\textbf{28.71} \cellcolor{"4-"}& 20.79 & 16.83 & 20.79 & \textbf{31.68} \cellcolor{"4-"}& 32.67 & 22.77 &38.61 & \textbf{44.55} \cellcolor{"4-"}\\
\midrule
Average & 12.58 & 14.30 & 14.37& \textbf{20.83} & 19.67 & 20.04 & 21.54 & \textbf{26.59} & 30.46 & 29.47 & 32.94 & \textbf{37.08} \\
\bottomrule
\end{tabular}
}
\end{table}

%% file: content/tab/conditional_results_CT.tex
\begin{table}[t]
\centering
\caption{\textbf{Conditional image retrieval results on CTRATE-IR.} Recall scores are in percentage. Anatomies are in descending order of their frequency in train set, with the `head' anatomies at the top and the `tail' anatomies at the bottom. The best results for each anatomy are bolded. Greener suggests higher improvement over baseline.}
\label{tab:conditioanl_results_ChestCT}
\resizebox{0.9\textwidth}{!}{%
\footnotesize  
\setlength{\tabcolsep}{4pt}
\begin{tabular}{llcccccc}
\toprule
\multirow{2}{*}{Anatomy} & \multirow{2}{*}{\#Samples} &
\multicolumn{2}{c}{$Recall@3 \uparrow$} & \multicolumn{2}{c}{$Recall@5 \uparrow$} & \multicolumn{2}{c}{$Recall@10 \uparrow$} \\
\cmidrule(lr){3-4} \cmidrule(lr){5-6} \cmidrule(lr){7-8}
 & & {CT-CLIP} & {RadIR} & {CT-CLIP} & {RadIR} & {CT-CLIP} & {RadIR} \\
\midrule
Bone                & 23.5k & 45.75 & \cellcolor{"0-7"} \textbf{49.76} & 56.33 & \textbf{60.51} \cellcolor{"0-7"} & 67.31 & \textbf{71.03}  \cellcolor{"0-7"}\\
Heart               & 23.3k & 33.75 & \cellcolor{"0-7"} \textbf{34.15} & 43.19 & \textbf{43.68} \cellcolor{"0-7"} & 55.72 & \textbf{59.44} \cellcolor{"0-7"}\\
Bronchie            & 21.7k & 55.18 & \cellcolor{"0-7"} \textbf{57.76} & 67.20 & \cellcolor{"0-7"}\textbf{69.42} & 75.81 & \textbf{78.43} \cellcolor{"0-7"} \\
Trachea             & 21.7k & 57.43 & \cellcolor{"0-7"}\textbf{60.48} & 69.24 & \cellcolor{"0-7"} \textbf{70.51} & 77.71 & \textbf{80.61} \cellcolor{"0-7"} \\
Pleura              & 18.2k & 35.14 & \cellcolor{"0-7"} \textbf{40.57} & 44.59 & \textbf{54.14} \cellcolor{"0-7"} & 60.00 & \textbf{71.64} \cellcolor{"7-14"}\\
Vertebrae           & 13.5k & 57.69 & \cellcolor{"0-7"} \textbf{62.04} & 63.69 & \textbf{66.91} \cellcolor{"0-7"} & 71.89 & \textbf{73.56} \cellcolor{"0-7"}\\
Liver               & 12.5k & 72.97 & \cellcolor{"0-7"} \textbf{78.14} & 77.58 & \textbf{79.26} \cellcolor{"0-7"} & 79.81 & \textbf{80.23} \cellcolor{"0-7"}\\
Aorta               & 11.8k & 48.90 & \cellcolor{"7-14"} \textbf{52.44} & 54.04 & \textbf{59.26} \cellcolor{"0-7"} & 62.56 & \textbf{65.93} \cellcolor{"0-7"}\\
Spinal canal        & 2.4k & 76.39 & \cellcolor{"0-7"}\textbf{79.17} & 83.33 & \textbf{90.28} \cellcolor{"0-7"}& 90.28 & \textbf{91.67} \cellcolor{"0-7"}\\
Gallbladder         & 2.4k & 19.10 & \cellcolor{"7-14"} \textbf{32.58} & 25.84 & \textbf{42.70} \cellcolor{"7-14"}& 39.33 & \textbf{52.81} \cellcolor{"7-14"}\\
Clavicle            & 1.2k & 64.29 & \cellcolor{"14-"} \textbf{89.29} & 75.00 & \textbf{96.43} \cellcolor{"14-"} & 96.43 & \textbf{100.00} \cellcolor{"0-7"}\\
Ascending aorta     & 1.6k & 23.73 & \cellcolor{"14-"} \textbf{48.28} & 37.29 & \textbf{56.90} \cellcolor{"14-"}& 50.85 & \textbf{65.52} \cellcolor{"14-"}\\
Pulmonary artery    & 1.6k & 18.18 & \cellcolor{"7-14"} \textbf{28.79} & 31.82 & \textbf{50.00} \cellcolor{"14-"} & 53.03 & \textbf{68.18} \cellcolor{"7-14"}\\
Breast               & 1.1k & 54.17 & \cellcolor{"14-"} \textbf{73.91} & 75.00 & \textbf{78.26} \cellcolor{"0-7"}& 75.00 & \textbf{91.30} \cellcolor{"14-"}\\
Pancreas             & 0.8k & 20.51 & \cellcolor{"14-"} \textbf{48.72} & 38.46 & \textbf{61.54} \cellcolor{"14-"} & 56.41 & \textbf{74.36} \cellcolor{"14-"}\\
Stomach              & 0.8k & 33.33 & \cellcolor{"14-"} \textbf{54.17} & 45.83 & \textbf{75.00} \cellcolor{"14-"}& 79.17 & \textbf{95.83} \cellcolor{"14-"}\\
\midrule
Average          &      /       & 43.85 & \textbf{55.23} & 54.44 & \textbf{66.29} & 67.09 & \textbf{76.12} \\
\bottomrule
\end{tabular}
}
\vspace{-.2cm}
\end{table}

%% file: content/06_Conclusion.tex
\section{Conclusion}
In this paper, we propose a novel methodology that leverages dense radiology reports to define image-wise similarity ordering at multiple granularities in a scalable and fully automatic way. We contribute two comprehensive datasets, MIMIC-IR and CTRATE-IR, with comprehensive and fine-grained image similarity ranking annotations for Chest X-ray and CT images. We build RadIR-CXR and RadIR-ChestCT, which demonstrate superior performance in diverse retrieval 
tasks, and could meet clinical demands flexibly by supporting fine-grained image retrieval conditioned on anatomy.


%% file: content/07_Appendix.tex
\appendix
\section{Appendix}

\begin{figure}[!h]
\centering
\includegraphics[width=0.95\textwidth]{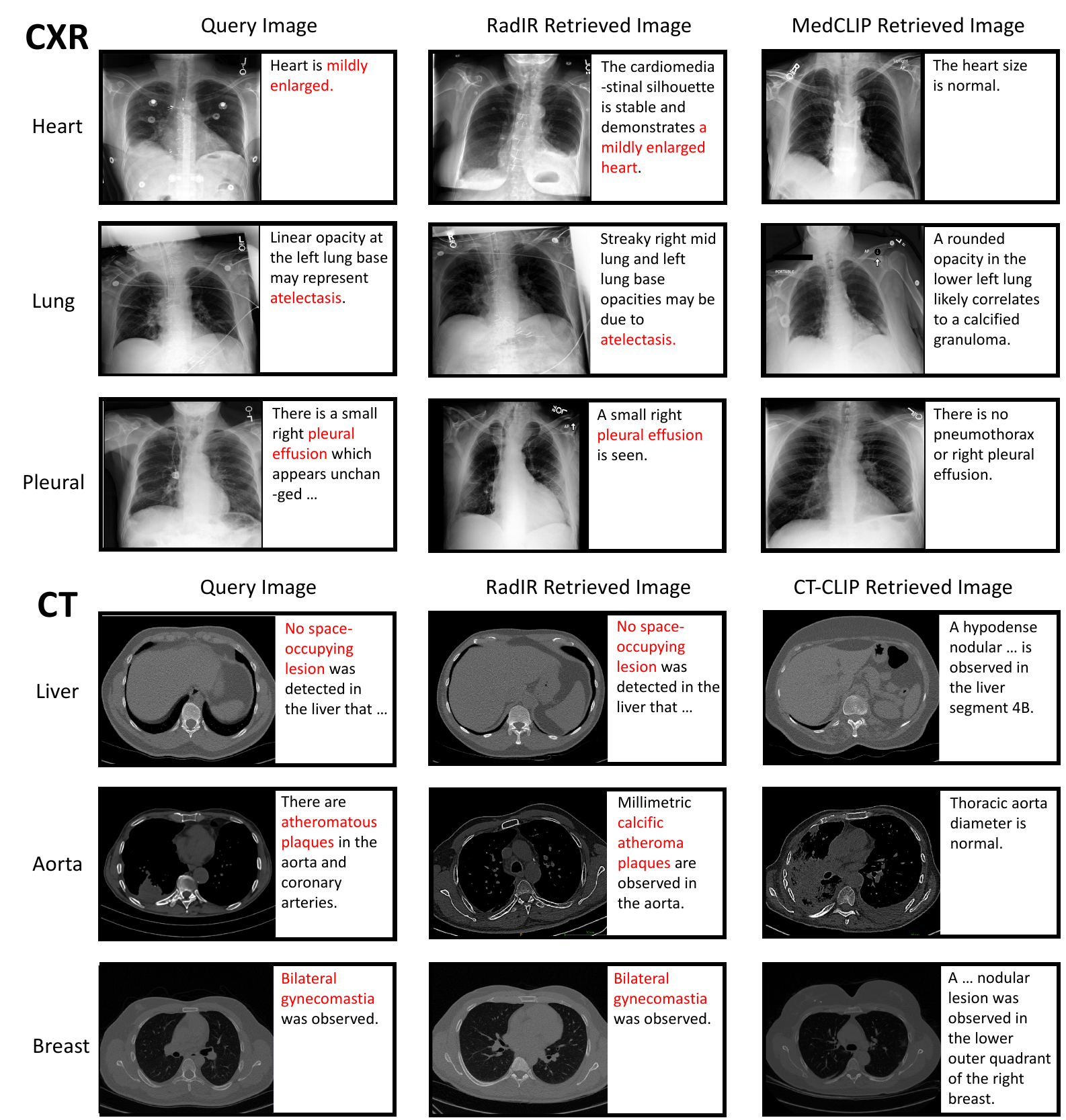}
\caption{\textbf{Qualitative comparison of conditional image retrieval results on MIMIC-IR and CTRATE-IR.} Heart, lung, pleural, liver, aorta and breast are used as query conditions respectively. The query image and the top 1 retrieved images from RadIR and the strongest baselines are presented, each with regional findings attached for reference.
}
\label{fig:visualization}
\end{figure}

